\documentclass[letterpaper]{article} 
\usepackage{aaai25}
\usepackage{times}  
\usepackage{helvet}  
\usepackage{courier}  
\usepackage[hyphens]{url}  
\usepackage{graphicx} 
\usepackage{natbib}  
\usepackage{caption} 
\usepackage{algorithm}
\usepackage{algorithmic}

\usepackage{newfloat}
\usepackage{listings}
\DeclareCaptionStyle{ruled}{labelfont=normalfont,labelsep=colon,strut=off} 
\floatstyle{ruled}
\newfloat{listing}{tb}{lst}{}
\floatname{listing}{Listing}
\title{Utilizing Large Language Models for Event Deconstruction to Enhance Multimodal Aspect-Based Sentiment Analysis}
\author{
    Xiaoyong Huang\textsuperscript{\rm 1},
    Heli Sun\textsuperscript{\rm 1},
    Qunshu Gao\textsuperscript{\rm 1},
    Wenjie Huang\textsuperscript{\rm 1},
    Ruichen Cao\textsuperscript{\rm 1}
}
\affiliations{
    \textsuperscript{\rm 1}Faculty of Electronic and Information Engineering, Xi'an Jiaotong University\\

    hxy{\_}computer@stu.xjtu.edu.cn
}

\usepackage{bibentry}

\begin{document}

\maketitle

\begin{abstract}
With the rapid development of the internet, the richness of User-Generated Contentcontinues to increase, making Multimodal Aspect-Based Sentiment Analysis (MABSA) a research hotspot. Existing studies have achieved certain results in MABSA, but they have not effectively addressed the analytical challenges in scenarios where multiple entities and sentiments coexist. This paper innovatively introduces Large Language Models (LLMs) for event decomposition and proposes a reinforcement learning framework for Multimodal Aspect-based Sentiment Analysis (MABSA-RL) framework. This framework decomposes the original text into a set of events using LLMs, reducing the complexity of analysis, introducing reinforcement learning to optimize model parameters. Experimental results show that MABSA-RL outperforms existing advanced methods on two benchmark datasets. This paper provides a new research perspective and method for multimodal aspect-level sentiment analysis. The related code will be open-sourced for further research.
\end{abstract}

%

\section{Introduction}

With the rapid development of the Internet, user-generated content has become increasingly rich. How to accurately mine users' emotional information from massive multimodal data has become a research hotspot in the field of multimodality. Sentiment analysis, as an important branch of data mining, aims to identify and analyze subjective emotional tendencies within texts. Among these, Multimodal Aspect-Based Sentiment Analysis (MABSA) focuses on analyzing users' emotional expressions towards a particular aspect or object with the assistance of image data, which holds high practical application value \cite{yang2024conditioned}.

There is already a lot of excellent work being done in the area of aspect-based sentiment analysis \cite{cao2022aspect}proposed an undirected differential emotion framework that eliminates affective biases to obtain stronger representations for sentiment classification \cite{zhang2022ssegcn} improved the accuracy of aspect-level sentiment analysis by learning semantic associations related to aspects and the global semantics of sentences through syntactic dependency trees. Considering multimodal input \cite{zhou2023aom},addressed the reduction of visual and textual noise brought about by complex image-text interactions.

\begin{figure}[t]
\centering
\includegraphics[width=0.45\textwidth]{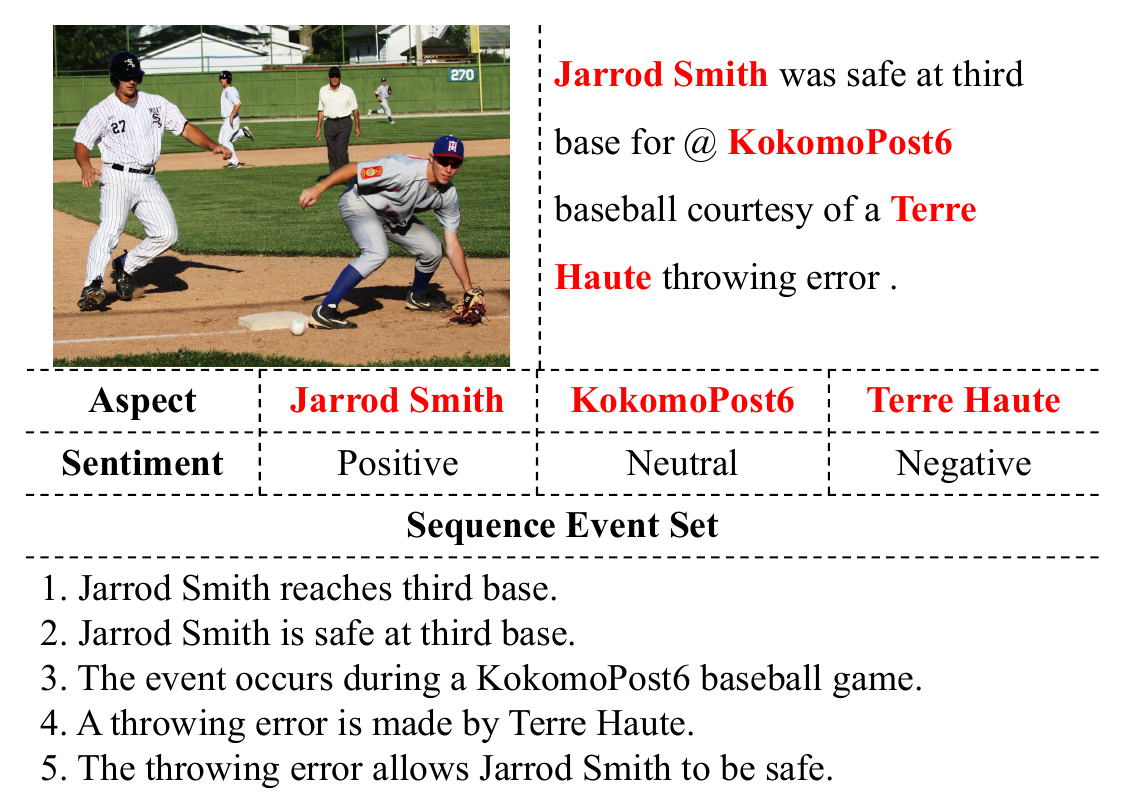} 
\caption{An example of Multimodal Aspect-Based Sentiment Analysis. It includes three aspects: Jarrod Smith, KokomoPost6,Terre Haute, along with their corresponding sentiments. In addition, we present a Sequence Event Set obtained through event decomposition by Qwen-Max-0428\cite{qwen1.5}.}
\label{fig2}
\end{figure}

However, we argue that the aforementioned work does not take into account the following challenge: the complexity of multimodal aspect-based sentiment analysis mainly stems from the fact that texts often contain multiple aspect terms, and each term may carry different sentiment polarities. As shown in Figure 1, the multimodal data contains three aspect terms, and the sentiments corresponding to these terms are all distinct. Traditional sentiment analysis models often struggle to achieve precise aspect term identification and sentiment prediction when confronted with scenarios featuring multiple aspect terms and coexisting sentiments.

To address this issue, our paper innovatively employs Large Language Models (LLMs) for event decomposition, refining the original text into sub-events that contain single or a few entities. LLMs such as ChatGPT and Qwen \cite{yang2024qwen2} have demonstrated remarkable capabilities across various natural language processing tasks, capable of extracting information from text via specific instructions, aiding in the construction of knowledge graphs among other tasks \cite{xu2023large}. Following this line of thinking, we utilize LLMs to decompose the original text into a set of events, where each sub-event contains only one or a few aspect terms, as illustrated by the Sequence Event Set in Figure 1. The advantage of this approach lies in the fact that each sub-event involves only one or two points of evaluation, significantly reducing the complexity of the sentiment analysis task. Moreover, since each sub-event in the event set can be considered in chronological order, the event set can be regarded as a Sequence Event Set. Given the superior performance of reinforcement learning in sequential tasks, we can incorporate it into the multimodal aspect-based sentiment analysis task to enhance the accuracy of aspect term prediction and sentiment analysis.

Specifically, we propose a reinforcement learning for Multimodal Aspect-based Sentiment Analysis (MABSA-RL). This framework initially breaks down the original text into a Sequence Event Set using a text decomposition module, extracting sub-events to reduce the complexity of aspect term prediction and sentiment analysis. Subsequently, we design a simple multimodal aspect prediction and sentiment analysis agent. We set up a specialized reinforcement learning environment based on the Sequence Event Set, pre-training the agent with supervised imitation learning and optimizing it with REINFORCE \cite{williams1992simple} reinforcement learning policy to improve model performance.

Our contributions are as follows:
\begin{itemize}
    \item We innovatively propose an event decomposition strategy based on LLMs, which refines the original text into a Sequence Event Set through specific instructions. Each sub-event in the Sequence Event Set contains only a single or a few aspect terms. This method effectively reduces the complexity of multimodal sentiment analysis, as each sub-event involves only one or two evaluation points, thereby simplifying the sentiment analysis process.
    \item Targeting the characteristics of the Sequence Event Set, we design a specialized reinforcement learning environment for the MABSA task. By pre-training with imitation learning and optimizing with the REINFORCE algorithm, we improve the strategies for aspect identification and sentiment prediction, enhancing the model's performance. To our knowledge, this is the first work that applies reinforcement learning to MABSA tasks.
    \item We develop a framework called MABSA-RL, which provides a new perspective on applying reinforcement learning to non-sequential decision-making tasks.
    \item Experiments on two benchmark datasets demonstrate that the MABSA-RL framework outperforms state-of-the-art methods overall. This validates the effectiveness of our approach in MABSA tasks. Furthermore, our code will be open-sourced to facilitate further exploration and validation of our method by other researchers.
\end{itemize}

\section{Related works}
\subsection{MABSA}
Previous work in multimodal aspect-based sentiment analysis has primarily focused on modal alignment. For instance, JML \cite{ju2021joint} developed an auxiliary text-image relationship detection module within a hierarchical framework to achieve multimodal integration. UMAEC \cite{2022-20441} established a shared feature module to capture semantic relationships between tasks. DTCA \cite{yu2022dual} enhanced inter-modal attention by introducing additional auxiliary tasks. VLP-MABSA \cite{ling-etal-2022-vision} transformed the analysis task into a text generation problem, reinforcing the model’s understanding of aspects, opinions, and their coherence through specific pretraining tasks. Recent trends have concentrated on strengthening sentiments and aspects. CMMT \cite{yang2022cross} learned intra-modal representations of sentiments and aspects via auxiliary tasks and introduced a text-guided cross-modal interaction module to modulate the contribution of visual information. GMP \cite{yang-etal-2023-shot-joint} predicted the number of aspects in instances through multimodal prompts. AESAL \cite{Zhu2024JointMA} constructed aspect-enhanced pretraining tasks and adopted a syntax-adaptive learning mechanism to discern differences in word importance within text. Atlantis \cite{xiao2024atlantis} augmented multimodal data by incorporating visual aesthetic attributes. FITE \cite{yang2022face} concentrated on capturing visual emotional cues through facial expressions, selectively matching and fusing them with textual modalities pertaining to target aspects.

Despite these successes, they overlooked the fundamental issue that the complexity of multimodal aspect-based sentiment analysis stems from the presence of multiple aspect terms in the text, each potentially bearing different sentiment polarities. To address this, we leverage LLMs to decompose texts into Sequence Event Set, where each sub-event contains only one to two aspect terms, thereby reducing the task's complexity.
\subsection{Reinforcement Learning}
Deep Reinforcement Learning (DRL), combining the powerful representation capabilities of deep learning with the decision optimization abilities of reinforcement learning, has achieved remarkable results across various domains such as games \cite{ye2020towards}, robotics control \cite{tang2024deep}, autonomous driving \cite{kiran2021deep}, and medical decision-making \cite{hao2022hierarchical}. Since the mathematical foundation and modeling tools of reinforcement learning are rooted in Markov Decision Processes, it has been predominantly applied to sequential decision-making tasks \cite{ladosz2022exploration}. 

Our proposed MABSA-RL framework utilizes LLMs to transform non-sequential decision tasks into sequential ones, enabling the application of reinforcement learning techniques to non-sequential decision problems, thus offering a novel approach for future research in handling such tasks.

\section{Methodology}
In this section, we first introduce the task formulation, followed by a detailed description of the proposed MABSA-RL framework. Figure 2 illustrates the overall architecture of MABSA-RL, which consists of a Text Decomposition Module, a Multimodal Aspect Prediction and Sentiment Analysis Agent, and a Sequential Decision Enhancement Module. Specifically, we first employ LLMs to decompose the text into a Sequence Event Set. Subsequently, an agent is designed to predict the probability distributions of aspect terms and sentiments using both textual and visual information. Finally, based on the Sequence Event Set, the non-sequential decision-making task is transformed into a sequential decision-making task. We utilize supervised cloning learning and the reinforcement learning algorithm REINFORCE to update the agent's parameters, thereby enhancing the quality of aspect term prediction and sentiment analysis.

\begin{figure*}[t]
\centering
\includegraphics[width=0.8\textwidth]{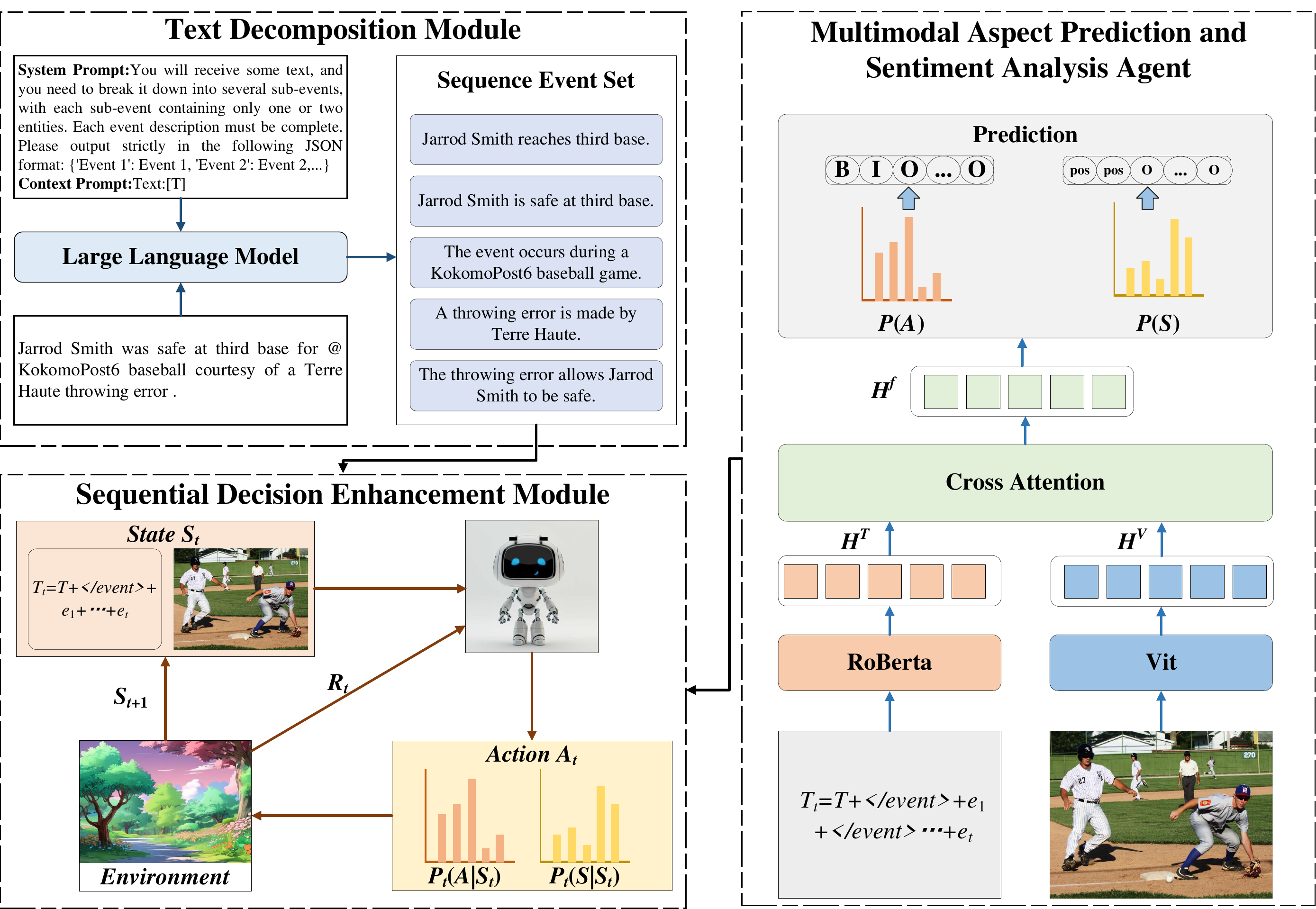} 
\caption{The overview of MABSA-RL.}
\label{fig2}
\end{figure*}

\subsection{Task Formulation}
Formally, we assume that the dataset ${D=\{(T_{i},V_{i},A_{i},S_{i})_{i=1}^{K}\}}$ 
consists of $K$ samples. For each sample $x\in D$ it includes a text ${T=\{t_1,t_2,\ldots,t_n\}}$ composed of n words, an associated image ${V\in R}^{3\times{H}\times{W}}$ , and aspects ${A=\{a_1,a_2,\ldots,a_m\}}$ consisting of m words along with their corresponding sentiments ${S=\{s_1,s_2,\ldots,s_m\}}$, where 3,$H$,$W$ denote the number of channels, height, and width of the image, respectively. $a_i$ denotes the \( i \)-{th} aspect item, and ${s_i\in\{POS,NEU,NEG\}}$ denotes the sentiment corresponding to the \( i \)-{th} aspect item, with POS, NEU, NEG representing positive, neutral, and negative sentiments, respectively. Our objective is to learn a model ${F(T,V)\to(A,S)}$ , that is, given $T$ and $V$, predict $A$ and $S$ .
\subsection{Text Decomposition Module}

As we introduced earlier, in MABSA tasks, texts often accompany multiple aspect items, each potentially bearing different sentiment polarities. To reduce the complexity of the MABSA task and improve model performance, as shown in Equation (1), we employ LLMs to decompose the original text into a set of events, where \( l \) represents the number of events in the set, and \( e_j \) is a sub-event containing a single or a few aspect items.\begin{equation}E=LLMs(T)\end{equation}
\par
We use the prompts listed in Table 1 to ensure that the narrative of each \( e_j \) is complete. Since each sub-event in the event set \( E \) is decomposed according to the narrative order from front to back in \( T \), each sub-event in \( E \) exhibits a certain temporal sequence. Based on this, \( E \) can be considered as a sequence, so we refer to \( E \) as a Sequence Event Set.

\begin{table}[h]
\centering
\begin{tabular}{l}
\hline
    \textbf{System Prompt:}You will receive some text, and you need \\to break it down into several sub-events, with each sub-\\event containing only one or two entities. Each event\\description must be complete. Please output strictly in the\\following JSON format: \{'Event 1': Event 1, 'Event 2': \\Event 2,...\}\\
\hline
    \textbf{Context Prompt:}Text:[T]\\
\hline

\end{tabular}
\caption{Text Decomposition Prompt}
\label{table1}
\end{table}

\subsection{Multimodal Aspect Prediction and Sentiment Analysis Agent}
We designed a straightforward multimodal aspect prediction and sentiment analysis agent. Specifically, we input a state \( S_t = \{T_t, V\} \), where \( T_t \) represents the textual information at time \( t \), and \( V \) denotes the image. We append two special tokens $[CLS]$ and $[SEP]$ at the beginning and end of the text as sentence start and end markers, and use $[CLS]$ as the marker for the start of the image. Then, we utilize RoBERTa \cite{liu2019roberta} to extract text embeddings and employ ViT \cite{dosovitskiy2020image} to extract visual embeddings from the image.
\begin{equation}H^T=\mathrm{RoBERTa}(T_t)\end{equation}
\begin{equation}H^V\mathrm{=MLP~(ViT(V))}\end{equation}
\par
Among these, we used an MLP (Multi-Layer Perceptron) to adjust the shape of the extracted visual embedding \( H_v \) to match that of the text. ${H^T,H^V\in R^{n^t\times d}}$ ,where ${n^t}$ indicates the number of words, and \( d \) represents the dimension of the hidden state.
\par Subsequently, as illustrated by Equation (4), we apply a cross-attention mechanism to fuse $H^T\mathrm{~and~}H^V$ , obtaining the fusion embedding
$H^f\in R^{\mathrm{n^t\times d}}$:
\begin{equation}H^f=\mathrm{Softmax}(\frac{H^T\mathrm{W_Q}\times(H^V\mathrm{W_K})^\mathrm{T}}{\sqrt{\mathrm{d}}})\cdot(H^T\mathrm{W_V})\end{equation} 
\\where $W_Q$,$W_K$,$W_V$ are learnable parameters.
\par Following this, we obtain the probability distributions of the text's aspects \( A \) and sentiments \( S \) according to Equations (5) and (6):
\begin{equation}{P(A)}=\mathrm{softmax}(\mathrm{W_A}H^f+\mathrm{b_A})\end{equation}
\begin{equation}{P(S)}=\mathrm{softmax}(\mathrm{W_S}H^f+\mathrm{b_S})\end{equation}
where ${W_A}$ and ${W_S}$ are the weight matrices of the aspect and sentiment prediction layers, respectively, and ${b_A}$ and ${b_S}$ are the corresponding bias vectors.

\subsection{Sequential Decision Enhancement Module}
Our approach is to incrementally incorporate sub-events from the Sequence Event Set $E$ into the original text $T$, and then calculate the F1 score for aspect term extraction and sentiment prediction with respect to $T$. This F1 score serves as a reward to optimize the parameters of the entire agent.
\textbf{Environment Setup:}Following reinforcement learning terminology, we introduce states, actions, and rewards.\par
\textbf{{State} $\mathbf{S_t}$:}The state at time step $t$ consists of the current text ${T_t}$ and the associated image $V$, denoted as ${S_t} = \{{T_t},{V}\}$, ${T_t} = {T_{t-1}}+$ $</event>+{e_t}$.Specifically,${S_0} = \{{T_0},{V}\}$, ${T_0} = {T}$.The $</event>$ serves as an identifier.
\par \textbf{{Action} $\mathbf{A_t}$:}The action space contains all possible distributions of sentiments and aspects. The Agent outputs the predicted distributions of aspects $P_t(A|S_t)$ and sentiments $P_t(S|S_t)$ based on the state $S_t$.
\par \textbf{{Reward} $\mathbf{R_t}$:}Defined as the F1 score predicted for $T$ at time step $t$. Specifically, we first convert the probability distribution into predicted labels, then calculate the confusion matrix, and subsequently compute the F1 score. If we denote this process using a function $f_{1}(\cdot)$ , then our reward function can be expressed as:
\begin{equation}{R_t=(}f_1({P_t(A|S_t)})+f_1({P_t(S|S_t)}))/2\end{equation}

\par\textbf{Policy Network Setup:}We utilize the Multimodal Aspect Prediction and Sentiment Analysis Agent as the policy network $\pi_\theta({A_t|S_t})$, denoted with parameters $\theta$ representing the policy network.
\\
\textbf{Pre-training with Clone Learning: }To enhance subsequent training efficiency and avoid excessive random exploration during the reinforcement learning phase, we extract clone learning pre-training from any state $S_t=\{T_t, V\}$ across all training data. Using cross-entropy loss as the objective function, for the prediction of aspects and sentiments, we can define the loss functions as per Equations (8) and (9), with the overall loss function defined by Equation (10).
\begin{equation}L_A=-\sum_{i=1}y_{A,i}log(p_{A,i})\end{equation}
\begin{equation}L_{S}=-\sum_{i=1}y_{S,i}log(p_{S,i})\end{equation}
\begin{equation}L=0.5\times L_A+0.5\times L_S\end{equation}
\par Where $y_{A,i}$ and $y_{S,i}$ are the true labels for their respective categories, and $p_{A,i}$ and $p_{S,i}$ are the probabilities predicted by the model.\\
\textbf{Reinforcement Learning:}We update the policy parameters $\theta$ of the Agent using the REINFORCE algorithm to maximize the long-term return ${G_t=\sum_{k=t}^l\gamma^{k-t}R_k}$ where $\gamma$ is the discount factor. The update rule for each data instance is given by Equation (11).
\begin{equation}\theta=\theta+\alpha\cdot\nabla_{\theta}\mathrm{log~\pi_{\theta}(A_{t}|S_{t})}\cdot G_{t}\end{equation}
where $\alpha$ is the learning rate. The entire algorithmic process is detailed in Table 2. 

\begin{table}[h]
\centering
\begin{tabular}{l}
\hline
    {The algorithmic procedure of MABSA-RL:}\\
\hline
1. Use LLMs to decompose T into E .\\
2. Initialize Agent parameters $\theta$.\\
3. Conduct supervised learning clone  training, optimizing\\ $\theta$ until convergence.\\
4. Begin the reinforcement learning loop:\\
\qquad·Draw the next event ${e_t}$ from E,updating the state ${S_t}$.\\
\qquad·Utilize the Agent to compute  ${P_t}$ ${(A|S_t)}$ and ${P_t}$ ${(S|S_t)}$ \\\qquad based on ${S_t}$, predicting A and S.\\
\qquad·Calculate the reward ${R_t}$.\\
\qquad·Update ${\theta}$ to maximize ${G_t}$.\\
5. Repeat step 4 until all events in \( E \) have been processed.
\\
\hline
\end{tabular}
\caption{The algorithmic procedure of MABSA-RL.}
\label{Table2}
\end{table}



\section{Experiments}

\par In this section, we will verify the performance of MABSA-RL through experiments. Experimental setups, comparative models, experimental results, ablation studies, and case analyses will all be introduced.
\subsection{Experimental Setup}

  \quad\textbf{Datasets}: We conduct experiments on two multimodal benchmark datasets, including Twitter-2015 and Twitter-2017\cite{hu2019open}. Table 3 provides statistics on the datasets. These two Twitter datasets separately collected user posts published on Twitter during the periods of 2014-2015 and 2016-2017.

\begin{table}[h]
\setlength{\tabcolsep}{4.5pt}
\begin{center}
\begin{tabular}{ccccccc}
\hline
 & \multicolumn{3}{c}{Twitter-2015} & \multicolumn{3}{c}{Twitter-2017} 
 \\
\cline{2-4}  
\cline{5-7}
                       & Train      & Dev      & Test     & Train      & Dev      & Test     \\
\hline
Positive               & 928        & 303      & 317      & 1508       & 515      & 493      \\
Neutral                & 1883       & 670      & 607      & 1638       & 517      & 573      \\
Negative               & 368        & 149      & 113      & 416        & 144      & 168      \\
Total Aspects         & 3179       & 1122     & 1037     & 3562       & 1176     & 1234     \\
Total Sentence        & 2101       & 727      & 674      & 1746       & 577      & 587    \\ 
\hline

\end{tabular}
\end{center}
\caption{Statistics of the two benchmark datasets. The first three rows represent the counts of each sentiment type across both datasets. Rows four and five indicate the number of aspect terms and sentences, respectively.}
\label{Table3}
\end{table}

\textbf{Hyperparameter Settings:} Our experiments were implemented under the PyTorch framework utilizing NVIDIA 3090 GPUs. The learning rate was set to 2e-5 during the supervised clone learning phase and adjusted to 1e-5 for the reinforcement learning stage. The dimension of the hidden layer was 768, with dropout set to 0.1.For the LLM, we utilize Qwen-Max-0428.
\par \textbf{Evaluation Metrics:} To assess the performance of the algorithms, in line with previous work, we utilize Micro-F1 (F1), Precision (P), and Recall (R) to evaluate our model. Higher metrics indicate superior model performance.
\subsection{Comparative Models}
We compare the proposed MABSA-RL against three textual Aspect-Based Sentiment Analysis(ABSA) methods and eight MABSA methods.
\par Methods for ABSA:

\par
${1)}$ \textbf{SPAN} \cite{hu2019open} directly extracts multiple opinion targets and identifies sentiment polarities from sentences under supervision that spans boundaries.${2)}$ \textbf{D-GCN} \cite{chen2020joint} models syntactic dependencies using GCN \cite{kipf2016semi}.
${3)}$ \textbf{BART} \cite{yan2021unified} is a pre-trained sequence-to-sequence model that addresses all ABSA subtasks within an end-to-end framework.
\par
Methods for MABSA:
\par
${1)}$ \textbf{UMT-collapse} \cite{yu2020improving}, OSCGA-collapse \cite{wu2020multimodal}, and rbert-collapse \cite{sun2021rpbert} use the same visual input to fold individual tokens.
${2)}$ \textbf{UMT+TomBERT}, \textbf{OSCGA+TomBERT} are two pipelined approaches combining UMT, OSCGA with TomBERT respectively.${3)}$ \textbf{JML} \cite{ju2021joint} is a multimodal joint method capable of handling aspect term extraction and sentiment classification simultaneously.${4)}$ \textbf{VLP-MABSA} \cite{ling-etal-2022-vision} is a unified multimodal encoder-decoder architecture for all pre-training and downstream tasks.${5)}$ \textbf{CMMT} \cite{yang2022cross} is a multitask learning framework for extracting aspect-sentiment pairs from pairs of sentences and images.${6)}$ \textbf{AOM} \cite{zhou2023aom} is an aspect-oriented network designed to alleviate the noise in vision and text produced by complex image-text interactions.${7)}$ \textbf{Atlantis} \cite{xiao2024atlantis} augments multimodal data by introducing visual aesthetic attributes.${8)}$ \textbf{AESAL}\cite{Zhu2024JointMA} designs a syntactic adap-
tive learning mechanism to capture the difference in the importance of different words in the text.

\subsection{Experimental Results}

\begin{table*}[t]
\centering
\begin{tabular}{ccccccccc}
\hline
{}                 & {Methods}         & {Venue} & \multicolumn{3}{c}{Twitter-2015}                                       & \multicolumn{3}{c}{Twitter-2017}                                       \\
\cline{4-6} \cline{7-9}
     &   & \multicolumn{1}{c}{}                       & \multicolumn{1}{c}{P} & \multicolumn{1}{c}{R} & \multicolumn{1}{c}{F1} & \multicolumn{1}{c}{P} & \multicolumn{1}{c}{R} & \multicolumn{1}{c}{F1} \\
\hline{}  & SPAN                     & ACL 2020               & 53.7      & 53.9      & 53.8     & 59.6      & 61.7      & 60.6     \\
                           {Text-based}  & D-GCN                    & COLING 2020            & 58.3      & 58.8      & 59.4     & 64.2      & 64.1      & 64.1     \\
                             & BART                     & ACL 2021               & 62.9      & 65        & 63.9     & 65.2      & 65.6      & 65.4     \\
\hline {} & UMT+TomBERT              & ACL 2020\;IJCAI 2019    & 58.4      & 61.3      & 59.8     & 62.3      & 62.4      & 62.4     \\
            {}                 & OSCGA+TomBERT            & ACM MM 2020\;IJCAI 2019 & 61.7      & 63.4      & 62.5     & 63.4      & 64.0      & 63.7     \\
                             & OSCGA-collapse           & ACM MM 2020            & 63.1      & 63.7      & 63.1     & 63.5      & 63.5      & 63.5     \\
                             & RpBERT-collapse          & AAAI 2021              & 49.3      & 46.9      & 48.0     & 57.0      & 55.4      & 56.2     \\
                             & UMT-collapse             & ACL 2020               & 61.0      & 60.4      & 61.6     & 60.8      & 60.0      & 61.7     \\
                         {Multimodal}    & JML                      & EMNLP 2021             & 65.0      & 63.2      & 64.1     & 66.5      & 65.5      & 66.0     \\
                             & VLP-MABSA                & ACL 2022               & 65.1      & 68.3      & 66.6     & 66.9      & 69.2      & 68.0     \\
                             & CMMT                     & IPM 2022               & 64.6      & 68.7      & 66.5     & 67.6      & 69.4      & 68.5     \\
                             & AoM                      & ACL 2023               & 67.9      & 69.3      & 68.6     & 68.4      & 71.0      & 69.7     \\
                             & Atlantis                 & Inf.Fusion 2024        & 65.6      & 69.2      & 67.3     & 68.6      & 70.3      & 69.4     \\
                             & AESAL                    & IJCAI 2024             & \underline{67.3}      & \underline{70.4}      & \underline{69.1}     & \underline{69.4}      & \textbf{74.8}      & \underline{72.0}     \\
                             & {MABSA-RL}                 & Ours                    & \textbf{70.3}      & \textbf{71.7}      & \textbf{71.0}     & \textbf{73.2}      & \underline{73.1}      & \textbf{73.1}    \\
\hline
\end{tabular}
\centering
\caption{Results of different models on the MABSA task. The best results are highlighted in bold,and underline indicates the\\ second-best result.The same below.}
\end{table*}

\par
Table 4 demonstrates the results of various models on the MABSA task. Firstly, our proposed MABSA-RL significantly outperforms all text-based models, indicating the effectiveness of multimodal information in ABSA tasks. Secondly, compared to the state-of-the-art AESAL model, MABSA-RL boosts the P, R, and F1 values by 3\%, 1.3\%, and 1.9\% respectively on the Twitter-2015 dataset. On the Twitter-2017 dataset, the P value increases by 3.8\%, and the F1 value improves by 1.1\%. The slightly lower R value might be due to the imbalanced distribution of sentiments in the training data, particularly in the Twitter-2017 training set, where the number of negative instances is far less than those of the other two sentiments, causing the model to be biased towards predicting the majority sentiment, thus resulting in a lower recall.

Overall, MABSA-RL also outperforms other multimodal models. This is because previous research has primarily focused on the utilization of image and text information but neglected the complexity of multimodal aspect-level sentiment analysis, which mainly stems from the presence of multiple aspect terms in the text, each possibly carrying different sentiment polarities. Our proposed MABSA-RL tackles this issue by decomposing the textual information into a Sequence Event Set via LLMs, where each sub-event contains only a small number of evaluation points, reducing the difficulty for aspect term prediction and sentiment analysis. Additionally, by employing clone learning and reinforcement learning, we optimize the entire model, enhancing its predictive and decision-making capabilities.

\subsection{Ablation Studies}
In this section, we investigate the impact of each module on the final performance. The results of the ablation experiments are shown in Table 5. We use the multimodal aspect prediction and sentiment analysis agent as a baseline, training solely on the raw text and image without the enhancement provided by the Sequence Event Set. This allows us to understand the contributions of each component, such as the text decomposition module and the reinforcement learning strategy. 

\begin{table}[h]
\begin{tabular}{cccclccc}
\hline{Method} & \multicolumn{3}{c}{Twitter-2015}              & \multicolumn{3}{c}{Twitter-2017}              \\
\cline{2-4}\cline{5-7}
                        & P             & R             & F1            & P             & R             & F1            \\
                        \hline
Agent                   & 67.5          & 68.7          & 68.0          & 69.0          & 69.7          & 69.4          \\
${+}$Events                
& {\underline{69.8}}
& {\underline{70.5}}   
& {\underline{70.1}}    & {\underline{72.6}}    & {\underline{72.9}}    & {\underline{72.7}}    \\
${+}$RF                     & \textbf{70.3} & \textbf{71.7} & \textbf{71.0} & \textbf{73.2} & \textbf{73.1} & \textbf{73.1}
\\
\hline
\end{tabular}
\caption{Ablation Study of Individual Modules.${"}$Agent${"}$ refers to the multimodal aspect prediction and sentiment analysis agent trained with cross-entropy loss on the original data. "Events" signifies the use of Sequence Event Set for supervised training.${"}$RL${"}$ denotes the application of reinforcement learning. }
\end{table}

It can be observed that after incorporating the Sequence Event Set \( E \) for supervised training, there are improvements across all metrics on both benchmark datasets. On the Twitter-2015 dataset, the P, R, and F1 values increase by 2.3\%, 1.8\%, and 1.3\% respectively. Meanwhile, on the Twitter-2017 dataset, the P, R, and F1 values rise by 3.6\%, 3.2\%, and 3.3\% respectively. This directly validates the efficacy of the event decomposition module. Moreover, it indicates that the Sequence Event Set facilitates the simplification of aspect term prediction and sentiment analysis, thereby enhancing model performance.

\begin{figure}[ht]
\centering
\includegraphics[width=0.45\textwidth]{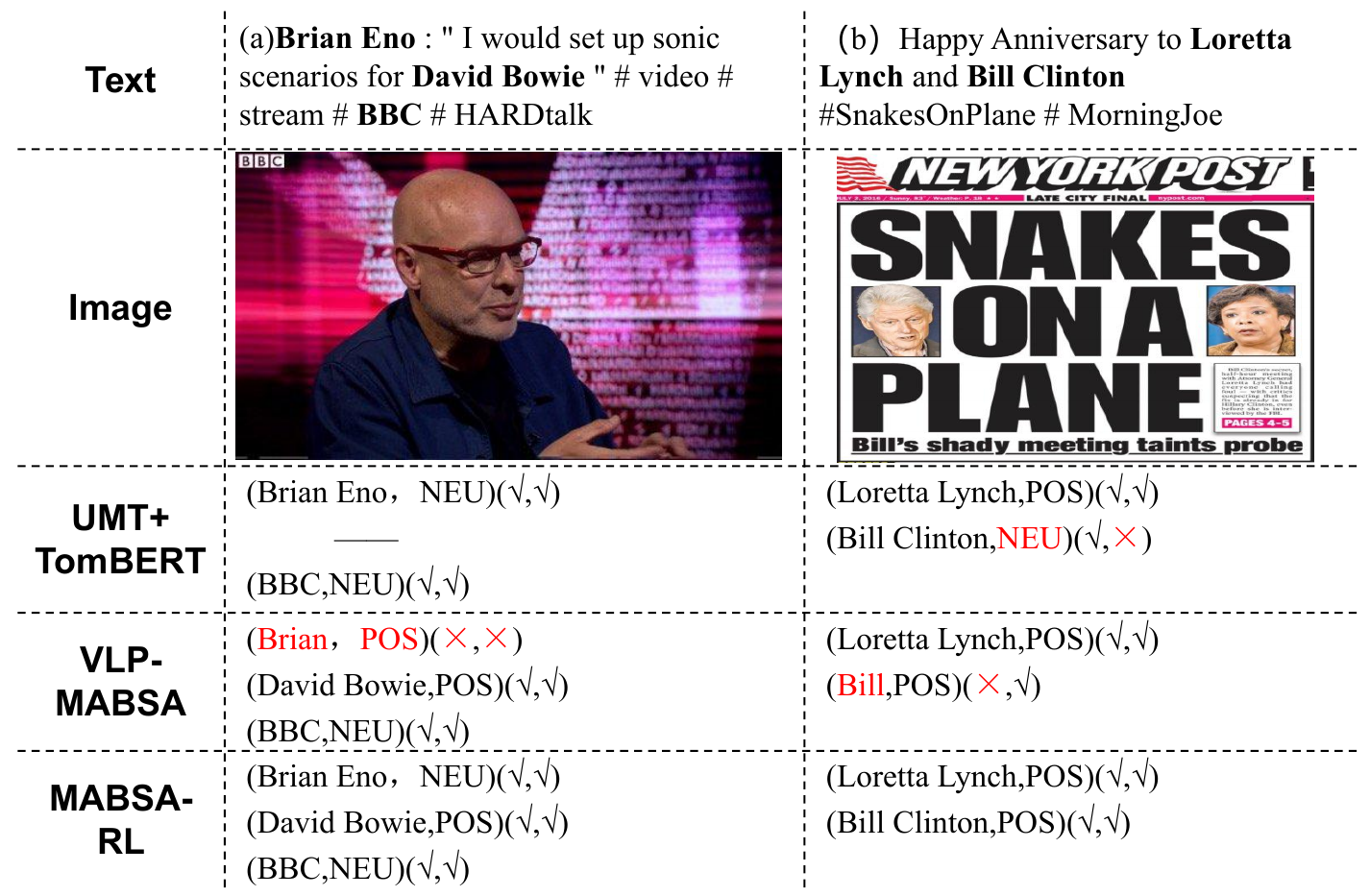} 
\caption{Two examples of predictions made by UMT+TomBERT, VLP-MABSA, and our MABSA-RL.}
\label{fig3}
\end{figure}
Upon the introduction of reinforcement learning, on the Twitter-2015 dataset, the P, R, and F1 values further increase by 0.5\%, 1.2\%, and 0.9\% respectively. Similarly, on the Twitter-2017 dataset, the P, R, and F1 values increment by 0.6\%, 0.2\%, and 0.4\% respectively. Clearly, the sequential decision-making training approach aids in boosting model performance. However, the performance gain attributed to reinforcement learning is relatively modest. We hypothesize that this is because the Sequence Event Set \( E \) can only be approximated as a sequence and does not perfectly align with the sequential nature required for reinforcement learning. Furthermore, inherent drawbacks of reinforcement learning, such as instability, have impacted the extent of performance improvement. These issues will be a focus in our future research endeavors.

\subsection{Case Study}

To further substantiate the effectiveness of MABSA-RL, we present a case study as follows. Figure 3 illustrates two examples of predictions made using UMT+TomBERT, VLP-MABSA, and our MABSA-RL. In Example (a), UMT+TomBERT failed to identify David Bowie and its corresponding sentiment. VLP-MABSA, on the other hand, did not fully recognize Brian Eno and incorrectly analyzed its sentiment. In Example (b), UMT+TomBERT misjudged the sentiment associated with Bill Clinton, while VLP-MABSA failed to completely recognize Bill Clinton. This may be due to the models' difficulties in analyzing the complex context under multiple aspect terms and varied sentiments, leading to incorrect aspect term predictions and sentiment judgments. In contrast, our proposed MABSA-RL correctly identified all aspect terms and provided accurate sentiment predictions in both cases. This is attributable to our use of LLMs to decompose textual information into a Sequence Event Set, where each sub-event contains only a small number of evaluation points, thereby reducing the complexity of aspect term prediction and sentiment analysis. Additionally, by employing reinforcement learning, we optimized the entire model, further enhancing its performance.

\section{Conclusion}
This paper proposes a reinforcement learning framework for multimodal aspect-level sentiment analysis called MABSA-RL. The framework encompasses a Text Decomposition Module, a Multimodal Aspect Prediction and Sentiment Analysis Agent, and a Sequential Decision Enhancement Module. Firstly, the Text Decomposition Module leverages LLMs to decompose text into a Sequence Event Set, with each sub-event containing only a limited number of appraisal points, thereby reducing the complexity for the model in predicting aspect terms and analyzing sentiments. Secondly, by constructing a Multimodal Aspect Prediction and Sentiment Analysis Agent, probability distributions for aspect term prediction and sentiment analysis are obtained. Lastly, within the Sequential Decision Enhancement Module, a specialized reinforcement learning environment is built for the Sequence Event Set, and the agent's parameters are optimized using behavior cloning and REINFORCE to enhance its performance. Experiments on two authoritative datasets demonstrate that MABSA-RL outperforms existing baseline methods in general, showcasing its superior performance. Furthermore, MABSA-RL offers new insights into applying reinforcement learning to non-sequential decision-making tasks.
\bigskip


\typeout{get arXiv to do 4 passes: Label(s) may have changed. Rerun}

\end{document}